%% file: root.tex
\documentclass[lettersize,journal]{IEEEtran}

\usepackage{amsmath}    
\usepackage{amssymb}    

\usepackage{mathtools}
\usepackage{mathrsfs}
\usepackage{pifont}
\usepackage{gensymb,soul}
\usepackage{graphics} 
\usepackage{subfigure}
\usepackage{booktabs}
\usepackage{epsfig} 
\usepackage{cite}
\usepackage{mathptmx} 
\usepackage{amsfonts,amssymb, amsmath}  
\usepackage[colorlinks,linkcolor=blue]{hyperref}

\usepackage{xcolor}
\usepackage{ifthen}
\usepackage{makecell}
\usepackage{algorithmicx}
\usepackage{algpseudocode}
\usepackage{threeparttable}
\usepackage[normalem]{ulem}

\usepackage[linesnumbered,ruled]{algorithm2e}

\DeclareMathAlphabet{\mathcal}{OMS}{cmsy}{m}{n}
\DeclareSymbolFont{largesymbols}{OMX}{cmex}{m}{n}

\usepackage{multirow}

\begin{document}

\title{ \LARGE \bf
GenHOI: Contact-Aware Humanoid-Object Interaction \\ by Imitating Generated Videos without Task-Specific Training

}

\author{
\IEEEauthorblockN{
Zhihai Bi$^{1}$, Qiang Zhang$^{2}$, Guoyang Zhao$^{1}$, Jiahang Cao$^{3}$,
Xueyin Luo$^{1}$, Yushan Zhang$^{1}$, Jinglan Xu$^{1}$,\\ Ruoyu Geng$^{1}$, Yulin Li$^{4}$, Andrew F. Luo$^{3}$, and Jun Ma$^{1,\dagger}$
}
\\
\IEEEauthorblockA{
\small
$^{1}$The Hong Kong University of Science and Technology (Guangzhou)
\quad \\
$^{2}$Artificial General Intelligence Institute, University of Science and Technology of China
\quad \\
$^{3}$The University of Hong Kong
\quad
$^{4}$National University of Singapore
\quad
$^{\dagger}$Corresponding author
}
}



\makeatletter
\newcommand{\TitleFig}[2]{
  \begingroup
  \def\@captype{figure}\caption{#1}\label{#2}%
  \endgroup
}
\makeatother

\IEEEaftertitletext{%
\vspace{-3.0em}
\begin{center}
\label{fig:cover}
\includegraphics[width=\linewidth]{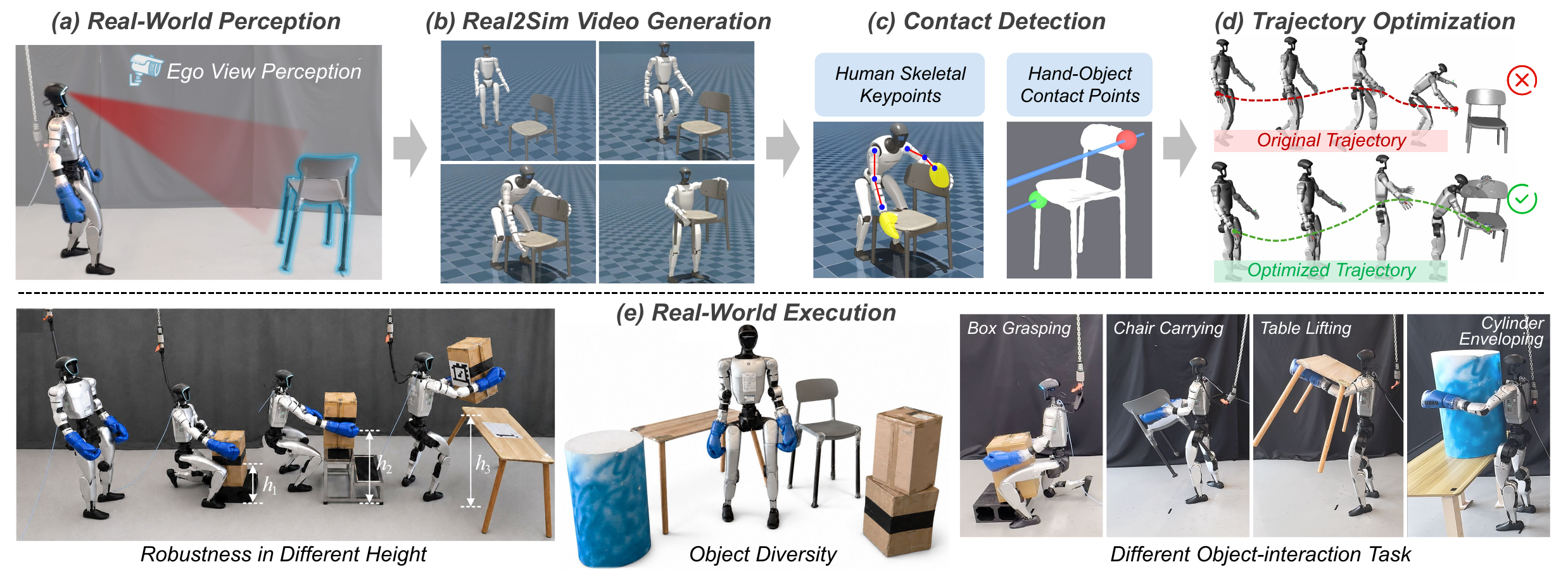}\TitleFig{%
\textbf{The proposed GenHOI framework enables a humanoid robot to perform diverse object-interaction tasks.}%
}{fig:cover}
\end{center}
\vspace{-2pt}%
}

\maketitle
\thispagestyle{empty}
\pagestyle{empty}


\begin{abstract}
Humanoid-Object Interaction (HOI) is essential for deploying humanoid robots in human-centered environments, yet it remains challenging due to the tight coupling between dynamic balance and stable interaction with diverse objects.
Existing methods often require time-consuming task-specific policy training or rely on rigid trajectory replay, which limits their ability to accommodate novel interaction scenarios.
In this work, we present \textit{GenHOI}, a simple yet effective framework that enables humanoid robots to perform diverse object-interaction tasks in a zero-shot manner by directly imitating a single generated video, without task-specific training or physical demonstration data. 
GenHOI first reconstructs the robot-object scene in simulation and renders a first-frame image, which conditions the synthesis of a task-oriented interaction video in conjunction with the language command.
The generated video is then analyzed to identify interaction-relevant contact events and estimate hand-object contact regions, which are encoded as object-centric geometric constraints that convert visual interaction cues into physically grounded optimization priors. Guided by these priors, the reference motion recovered from the video is refined and smoothed to resolve the scale ambiguity inherent in 2D video generation, while adapting a single reference trajectory to unseen robot-object relative poses. The optimized trajectory is finally executed by a closed-loop tracking controller.
We validate the proposed framework in extensive simulation and real-world experiments across diverse object-interaction tasks, including box grasping, asymmetric bimanual chair carrying, table lifting from below, and cylindrical-object enveloping. The project page is available at \url{https://genhoi-humanoid.github.io/}.
\end{abstract}

\section{Introduction} \label{sec:intro}
Humanoid robots hold great promise for operating in human-centered environments~\cite{gu2026humanoid}. Despite recent progress in locomotion, versatile Humanoid-Object Interaction (HOI) remains challenging because physical interaction with objects is tightly coupled with whole-body stability. The robot must simultaneously coordinate its motion, contact locations, and interaction forces across diverse objects~\cite{he2025viral}.

Current HOI methods either directly mimic a single reference trajectory~\cite{yang2025omniretarget} or learn task policies regularized by motion-distribution priors, such as motion-prior models~\cite{zheng2025embracing} and AMP-based discriminators~\cite{peng2021amp}. Essentially, these methods largely ground interaction learning in pre-collected motion data and therefore entail task-specific demonstrations, reward design, and policy training for each new task or object category. Moreover, since the behaviors are tied to reference motions, they often fail to generalize when the robot-object relative pose, object geometry, or contact affordance shifts beyond the training distribution \cite{lin2026prohoi}.

Recently, video generation models have made remarkable progress. Representative models such as Sora \cite{openai2024video} and Seedance \cite{seedance2026seedance} are capable of synthesizing videos that increasingly resemble the visual appearance and motion dynamics of the real physical world. Compared with existing model-based or data-driven methods for HOI, planning future actions through video generation offers a potentially more generalizable alternative.
Motivated by this observation, we explore a new paradigm for humanoid loco-manipulation: \textbf{Can a humanoid robot leverage the generalization capability of video generation models to perform loco-manipulation tasks, such as chair carrying, by imitating a single generated video?}


Recent efforts that leverage generated videos for robot manipulation can be broadly grouped into two lines of work. The first uses video generation models as auxiliary tools for policy learning, such as augmenting training data \cite{xie2026grail} or learning video-based representations \cite{liang2025video}. Although effective in improving generalization, these methods still depend on task-specific policy training and real robot interaction data. 
The second line of work seeks to directly translate generated videos into robot behaviors, avoiding the need for additional policy learning \cite{dharmarajan2025dream2flow}. However, existing demonstrations are mainly limited to fixed-base manipulators, where the workspace is constrained, and the robot-object interaction is relatively simple \cite{patel2025robotic}.
In contrast, extending direct video imitation to humanoid loco-manipulation is substantially more challenging, requiring the robot to coordinate whole-body locomotion with bimanual manipulation while establishing stable contacts with geometrically complex objects. In addition, spatial and metric inconsistencies between generated 2D videos and the real environment can lead to inaccurate contact localization and motion execution errors, a limitation largely overlooked by prior work \cite{zhou2026exoactor, xu2026morphology}. Moreover, existing paradigms are often tied to the scene configuration depicted in the video and therefore cannot generalize to unseen object poses. Comparisons with related approaches are presented in Table \ref{tab:related_work}.

\begin{table}[t]
    \centering
    \caption{Comparison with Related Works}
    \label{tab:related_work}
    \begin{tabular}{cccc}
        \toprule
        \textbf{Method} &
        \makecell{No Task-Specific \\ Training} &
        \makecell{Onboard \\ Sensing} &
        \makecell{Steerable \\ Interaction} \\
        \midrule
        HDMI \cite{weng2025hdmi}   & \textcolor{blue}{\ding{55}}  & \textcolor{blue}{\ding{55}} & \textcolor{blue}{\ding{55}}  \\
       
        Pro-HOI \cite{lin2026prohoi}  & \textcolor{blue}{\ding{55}}  & \textcolor{green}{\ding{51}} & \textcolor{green}{\ding{51}}\\

        ExoActor \cite{zhou2026exoactor}  & \textcolor{green}{\ding{51}} & \textcolor{blue}{\ding{55}} & \textcolor{blue}{\ding{55}} \\

        Dream2Act \cite{xu2026morphology}  & \textcolor{green}{\ding{51}} & \textcolor{blue}{\ding{55}} & \textcolor{blue}{\ding{55}} \\
        
        \textbf{Ours} & \textcolor{green}{\ding{51}}  & \textcolor{green}{\ding{51}} & \textcolor{green}{\ding{51}} \\
        \bottomrule
    \end{tabular}
\end{table}

To address these limitations, we propose GenHOI, a loco-manipulation framework for humanoid robots, as shown in Fig. \ref{fig:2}. The framework consists of four key modules: real-to-sim video generation, contact-aware geometric constraint extraction, geometry-guided trajectory optimization, and closed-loop trajectory tracking. First, the real-to-sim video generation module reconstructs the robot and target object in simulation and uses a simulator-rendered first-frame image, together with a language command, to condition a video diffusion model for synthesizing the video without relying on external cameras. Second, to mitigate the spatial and metric inconsistencies of the motion recovered from the 2D video, we derive contact-aware geometric constraints from the generated video and use them to guide trajectory optimization, thereby aligning the trajectory with the real robot workspace. Notably, this decoupled design allows a single trajectory to generalize across different robot-object relative poses by updating only the optimization constraints. Finally, the optimized trajectory is executed through closed-loop trajectory tracking with a general-purpose motion controller \cite{luo2025sonic}.
The main contributions are summarized as follows:
\begin{itemize}
    \item We propose an HOI framework that enables a humanoid robot to perform diverse object-interaction tasks in a zero-shot manner by directly imitating a single generated video, without requiring task-specific policy training or physical demonstration data.
    \item We introduce a contact-aware geometric constraint extraction module that identifies contact events in generated videos and estimates hand-object contact regions to construct object-centric geometric constraints, thereby converting visual cues into physically grounded priors.
    \item We develop a geometry-guided trajectory optimization pipeline that uses the extracted constraints to refine and smooth the reference motion, which mitigates metric discrepancies and enables a single reference trajectory to generalize to unseen robot-object relative poses.
    \item We validate the proposed framework through extensive simulation and real-world experiments. The results show that our method enables humanoid robots to learn diverse object-interaction behaviors more efficiently than the baselines while achieving stronger generalization.
\end{itemize}

\section{Related Work}\label{sec:related_work}

       


        

\subsection{Humanoid Loco-Manipulation}
In recent years, humanoid loco-manipulation has attracted increasing attention.
Some studies are inspired by the success of reinforcement learning (RL) in robotic locomotion and extend RL-based approaches to humanoid loco-manipulation tasks \cite{wang2025physhsi}. For example, DoorMan \cite{xue2025opening} designs a multi-stage reward function to train a policy capable of performing diverse articulated-object interaction tasks using only RGB perception. Similarly, VIRAL \cite{he2025viral} leverages reinforcement learning to achieve pick-and-place tasks. However, these approaches typically rely on task-specific reward engineering. In particular, for long-horizon, contact-rich tasks, training RL policies from scratch often requires curriculum learning, multi-stage setups, and extensive hyperparameter tuning.

Compared with training RL policies from scratch, a more efficient paradigm leverages offline human-object interaction demonstrations. These demonstrations are typically retargeted to the humanoid robot and then used to train a tracking policy \cite{shi2026egohumanoid}. For example, HDMI \cite{weng2025hdmi} extracts human and object motion trajectories from videos and trains an RL-based policy for interactions. The policy performs tasks such as door traversal robustly. Similarly, OmniRetarget \cite{yang2025omniretarget} introduces a mesh-based retargeting method that generates higher-quality interaction trajectories, enabling more challenging tasks such as fast platform climbing. Despite their effectiveness on specific tasks, they rely on carefully engineered preprocessing pipelines, including manually predefined robot-object contact points. Moreover, because they mainly learn to track fixed demonstration trajectories, they tend to generalize poorly to changes in robot-object relative pose.

Recently, inspired by the success of LLM and VLM, researchers explore more scalable learning paradigms, aiming to improve generalization through large-scale datasets \cite{zitkovich2023rt, o2023open}. Representative examples include EgoScale \cite{zheng2026egoscale}, which trains a VLA model on tens of thousands of hours of human egocentric videos, and $\Psi_0$ \cite{wei2026psi0}, which pretrains a VLM backbone on large-scale egocentric videos and further adapts it using high-quality real-world humanoid robot data. By combining broad pretraining on human video data with robot-specific post-training, these methods exhibit improved potential for cross-task transfer. However, their reliance on massive datasets and substantial computational resources limits their practicality for follow-up research.

In contrast to these paradigms, we explore a novel framework for humanoid loco-manipulation: directly imitating actions from generated videos, without relying on physical demonstrations or task-specific policy training.

\subsection{Video Generation for Robotics}
Recent advances in video generation models have enabled the synthesis of high-fidelity, controllable videos of robotic actions, offering a new technical pathway for robot learning \cite{zhou2024robodreamer}.
A common use of such models is data synthesis. DexImit \cite{mu2026deximit}, for instance, synthesizes bimanual dexterous manipulation sequences and converts them into training data for policy learning. DreamGen \cite{jang2025dreamgen} instead fine-tunes a video world model on a target robot, using it to generate synthetic data for downstream visuomotor policy training. Compared with manually collected demonstrations, such approaches can substantially reduce the cost of data acquisition.

Beyond data synthesis, recent work directly uses video generation models as policy backbones, jointly modeling future video frames and action sequences conditioned on language instructions and initial observations \cite{ma2026dit4dit}. For example, DreamZero \cite{ye2026world} builds on a pretrained video diffusion model to learn diverse skills from heterogeneous robot data. Unlike data-synthesis methods, this paradigm offers a more direct route from generative modeling to policy learning. However, the large model size and high inference latency of existing approaches limit their evaluation primarily to fixed-base manipulation, leaving their applicability to real-time humanoid loco-manipulation largely unexplored.

Another emerging direction seeks to recover executable control signals directly from generated videos without additional policy training. RIGVid \cite{patel2025robotic} guides manipulation by estimating the target-object pose in each video frame and tracking the extracted trajectory with the end effector under an initial grasp. Dream2Act \cite{xu2026morphology} instead extracts 3D robot coordinates from generated videos for tracking control. However, it assumes geometric consistency between the generated video and the real scene, leaving stable contact formation unresolved. Moreover, these methods rely on manually captured images as input and cannot be deployed in closed-loop settings using only onboard robot sensing.

To bridge these gaps, we reconstruct the robot and target object in simulation and use a simulator-rendered image for video generation. Furthermore, we extract contact-aware geometric constraints from the generated video to guide trajectory optimization. This design reduces spatial and metric discrepancies during real-world execution while enabling generalization across varying robot-object relative poses.

\section{Methodology}\label{sec:method}
\begin{figure*}[t]	
	\centering
	\includegraphics[width=0.98\linewidth]{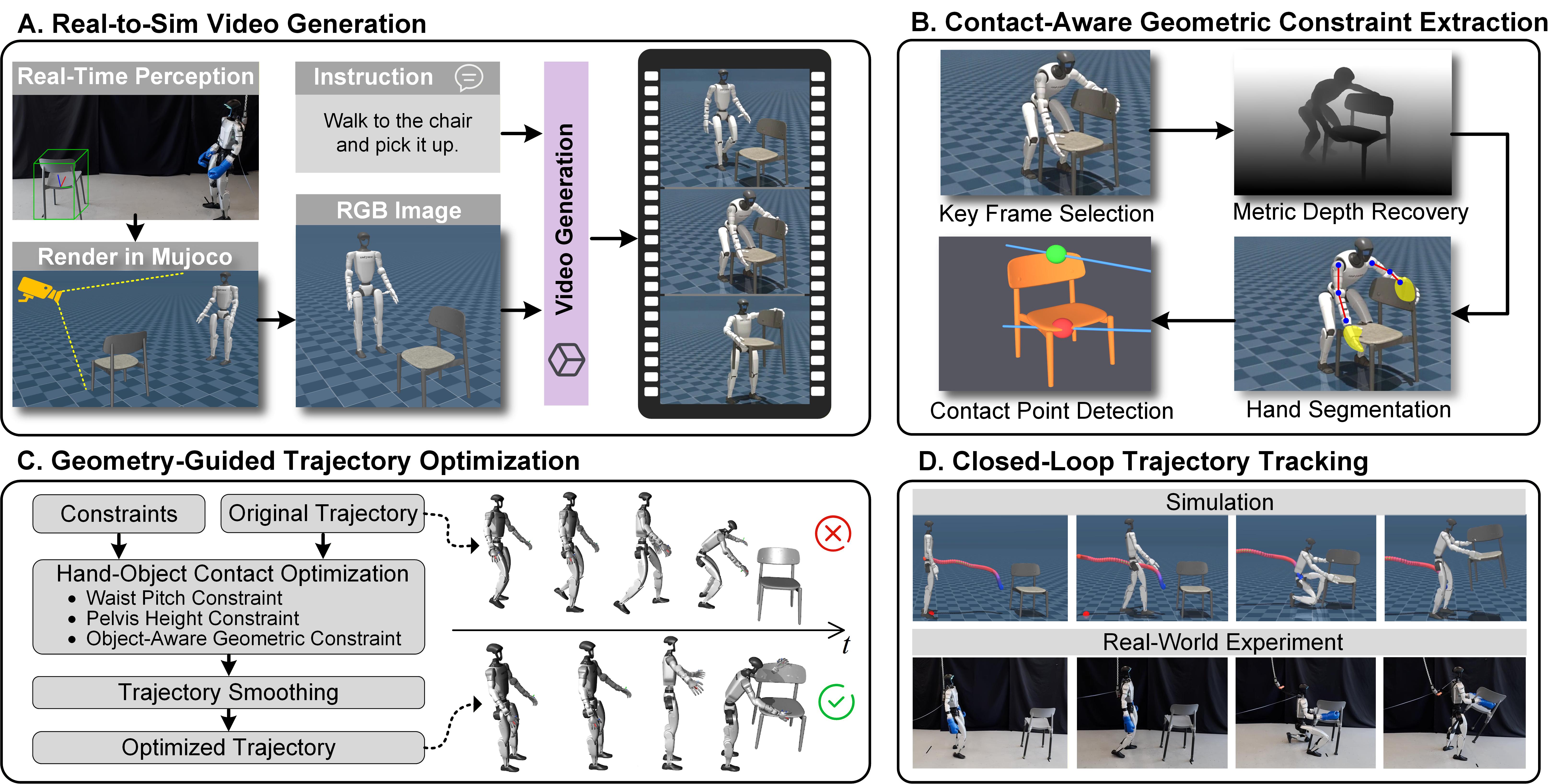}
	\setlength{\abovecaptionskip}{-1pt} 
	\caption{\textbf{Overview of the proposed GenHOI framework.} GenHOI first reconstructs the robot-observed scene in simulation and generates a reference video from the rendered image and text instructions (A). It then extracts contact-aware geometric constraints from the generated video via key-frame selection, metric depth recovery, hand segmentation, and contact point detection (B). These constraints are incorporated into trajectory optimization to refine the original trajectory into a physically feasible hand-object interaction trajectory (C), which is finally executed through closed-loop tracking in both simulation and real-world experiments (D).
    }
	\label{fig:2}
\end{figure*}

In this work, we propose GenHOI, a humanoid loco-manipulation framework that directly imitates generated videos without task-specific training, as illustrated in Fig. \ref{fig:2}. Specifically, the proposed framework comprises four key stages: real-to-sim video generation (Sec. \ref{sec:gen_video}), contact-aware geometric constraint extraction (Sec. \ref{sec:hand_obj_con}), geometry-guided trajectory optimization (Sec. \ref{sec:traj_opti}), and closed-loop trajectory tracking (Sec. \ref{sec:tracking}).

\subsection{Real-to-Sim Video Generation}\label{sec:gen_video}
In contrast to prior works that rely on external cameras to capture third-person images for video generation \cite{zhou2026exoactor}, our goal is to enable the robot to generate such videos using only onboard sensors, thereby supporting autonomous operation. To this end, we adopt a simple yet effective digital-twin-based solution, as shown in Fig. \ref{fig:2}(A). Specifically, assuming the object mesh is available, we first estimate the 6D pose of the object relative to the robot using onboard RGB-D observations based on foundationPose \cite{wen2024foundationpose} or AprilTag. The robot and the object are then rendered in a MuJoCo simulation environment. Thanks to the flexibility of the simulator, virtual cameras can be placed arbitrarily to obtain third-person views of the robot-object interaction. Given the simulated first-frame image and the corresponding language command, we then use Seedance 2.0 to generate a video (5\,s) from a fixed viewpoint (for details see Appendix \ref{appendix1}). 

Given the generated video with $N$ frames, we then recover the robot’s 3D pose sequence by following a standard pipeline widely adopted in prior work \cite{wang2026humanx}. Specifically, we first use GVHMR \cite{shen2024world} to estimate the 3D pose sequence from the input video, and then retarget it to the humanoid robot using GMR \cite{araujo2025retargeting}, yielding the raw robot 3D pose trajectory:
\begin{equation}
    \mathbf{h}_i= (\mathbf{h}_i^{\text{root}}, \mathbf{h}_i^{\text{joint}}),\quad i = 1,\cdots,N,
\end{equation}
where the $\mathbf{h}_i^{\text{root}} \in  \mathbb{R}^{6}$ indicates the 6D pose of the robot root, and  $\mathbf{h}_i^{\text{joint}} \in  \mathbb{R}^{29}$ denotes the robot joint angles.

\subsection{Contact-Aware Geometric Constraint Extraction}
\label{sec:hand_obj_con}

Due to depth ambiguity, occlusion, and perspective effects in monocular videos, the raw 3D pose trajectory recovered from generated videos often exhibits substantial inaccuracies in the global root motion and hand trajectories. Such errors can lead to physically inconsistent robot-object interactions during execution. To improve the feasibility of interaction, our core idea is to refine the raw 3D pose trajectory using hand-object contact poses.

To this end, we first recover accurate 3D hand positions from monocular videos. We uniformly sample frames from the last $N$ seconds of the generated video at 0.5\,s intervals and concatenate them in temporal order into a single composite image, which is fed into Doubao-Seed-2.0, a VLM model, to identify the key frame where the humanoid first establishes physical interaction with the object. Here, $N$ defines the temporal search window. We set $N=3$, which covers the typical contact interval while keeping the candidate set compact for efficient VLM-based key-frame selection (for details see Appendix \ref{appendix2}). This frame is referred to as the terminal interaction frame in the subsequent optimization. We then estimate its monocular depth using Depth Anything \cite{yang2024depth}. Since the predicted depth is defined only up to an unknown scale, we recover its metric scale using the digital-twin setup, where the camera parameters and object pose in the camera frame are available.

With the restored metric depth, we next use VitPose \cite{xu2022vitpose} to extract 2D keypoints for the left and right hands and lift them to coarse 3D positions in the world frame. However, due to errors in depth restoration and potential hand occlusions, these coarse hand positions may deviate from the true hand locations and therefore require further refinement. If the hand is visible and not occluded by the object, we refine the coarse hand position to the nearest point on the object mesh as the target contact position. Otherwise, we cast a ray from the camera center through the 3D hand position and use its last intersection with the object mesh as the final contact position. The hand visibility can be determined based on the color region within the hand mask. This process yields object-consistent target hand positions
\(\mathbf{p}_L^{\star}, \mathbf{p}_R^{\star} \in \mathbb{R}^3\)
in the world frame.

Since monocular depth and 2D keypoints do not provide reliable hand orientations, we retain the hand orientations from the recovered humanoid motion at the terminal interaction frame. Let
\(\mathbf{R}_L^{\star}, \mathbf{R}_R^{\star} \in SO(3)\)
denote the corresponding left and right hand orientations. Together with the target hand positions, these quantities define the contact-aware geometric constraints used in the subsequent optimization.

\subsection{Geometry-Guided Trajectory Optimization}
\label{sec:traj_opti}

Given a raw robot 3D pose trajectory \(\mathbf{h}_i, i=1,\cdots,N\), together with the extracted contact-aware terminal hand pose constraints, we refine the trajectory such that the resulting motion becomes more consistent with the desired robot-object interaction.
Directly optimizing the full-body humanoid motion would require jointly considering gait planning, foot contact transitions, and balance constraints, resulting in a highly coupled and computationally challenging problem~\cite{dafarra2022dynamic}. Instead, we perform a lightweight terminal refinement over the root pose and upper-body joints.

Specifically, we optimize a subset of the terminal full-body state:
\begin{equation}
\hat{\mathbf{h}}_N^{\mathrm{sub}}
=
\begin{bmatrix}
x & y & z & \psi & \phi & \mathbf{q}^{\top}
\end{bmatrix}^{\top},
\label{eq:substate}
\end{equation}
where \((x,y,z)\) denotes the root position, \(\psi\) denotes the root yaw angle, \(\phi\) denotes the waist pitch angle, and \(\mathbf{q}\in\mathbb{R}^{14}\) contains the upper-body arm joint angles.
Given \(\hat{\mathbf{h}}_N^{\mathrm{sub}}\), the world-frame poses of the left and right hands are computed through forward kinematics:
\begin{equation}
\begin{aligned}
\mathbf{T}_L(\hat{\mathbf{h}}_N^{\mathrm{sub}})
&=
\mathbf{T}_{\mathrm{root}}(x,y,z,\psi)\,
\mathbf{T}_{\mathrm{root}\rightarrow L}(\phi,\mathbf{q}_L),\\
\mathbf{T}_R(\hat{\mathbf{h}}_N^{\mathrm{sub}})
&=
\mathbf{T}_{\mathrm{root}}(x,y,z,\psi)\,
\mathbf{T}_{\mathrm{root}\rightarrow R}(\phi,\mathbf{q}_R),
\end{aligned}
\label{eq:hand_fk}
\end{equation}
where \(\mathbf{T}_{\mathrm{root}}\in SE(3)\) is the homogeneous transformation matrix of the robot root, and
\(\mathbf{T}_{\mathrm{root}\rightarrow L}(\cdot)\) and
\(\mathbf{T}_{\mathrm{root}\rightarrow R}(\cdot)\)
denote the forward kinematic mappings from the root frame to the left and right hand frames, respectively.

For a current pose \(\mathbf{T}=(\mathbf{R},\mathbf{p})\) and a target pose \(\mathbf{T}^{\star}=(\mathbf{R}^{\star},\mathbf{p}^{\star})\), we define the 6D pose error as:
\begin{equation}
\mathbf{e}(\mathbf{T},\mathbf{T}^{\star})
=
\begin{bmatrix}
\mathbf{p}-\mathbf{p}^{\star}\\
\mathrm{Log}\!\left((\mathbf{R}^{\star})^{\top}\mathbf{R}\right)
\end{bmatrix}
\in \mathbb{R}^{6},
\label{eq:pose_error}
\end{equation}
where the first three dimensions correspond to the translational error, and the last three dimensions represent the rotational error in the tangent space of \(so(3)\).

For bimanual object-holding motions, directly aligning the hands with the estimated object surface may lead to marginal contact under position-controlled execution. To encourage a stable holding interaction, we introduce a compliance-aware inward bias inspired by a virtual spring model. Let \(\mathbf{p}_L^{\star}\) and \(\mathbf{p}_R^{\star}\) be the translational components of \(\mathbf{T}_L^{\star}\) and \(\mathbf{T}_R^{\star}\), respectively. We define the inter-hand direction as:
\begin{equation}
\mathbf{d}
=
\frac{
\mathbf{p}_R^{\star}-\mathbf{p}_L^{\star}
}{
\left\|
\mathbf{p}_R^{\star}-\mathbf{p}_L^{\star}
\right\|_2
}.
\label{eq:inter_hand_direction}
\end{equation}
The target hand positions are then slightly displaced toward each other:
\begin{equation}
\bar{\mathbf{p}}_L^{\star}
=
\mathbf{p}_L^{\star}
+
\delta\mathbf{d},
\qquad
\bar{\mathbf{p}}_R^{\star}
=
\mathbf{p}_R^{\star}
-
\delta\mathbf{d},
\label{eq:inward_bias}
\end{equation}
where \(\delta>0\) is a small inward displacement. This inward displacement acts as a virtual compression between the two hands and induces a nominal holding tendency proportional to the imposed displacement, following the virtual spring relation \(F=k\delta\). Since the video does not provide reliable force information, this term is used as a heuristic compliance prior rather than an explicit contact-force estimate.
The adjusted terminal hand pose targets are defined as:
\begin{equation}
\bar{\mathbf{T}}_e^{\star}
=
\begin{bmatrix}
\mathbf{R}_e^{\star} & \bar{\mathbf{p}}_e^{\star} \\
\mathbf{0}^{\top} & 1
\end{bmatrix},
\qquad e\in\{L,R\}.
\label{eq:adjusted_hand_target}
\end{equation}
Using these targets, we define the geometric residual as:
\begin{equation}
\mathbf{r}_{\mathrm{geom}}(\hat{\mathbf{h}}_N^{\mathrm{sub}})
=
\begin{bmatrix}
\mathbf{W}_L^{1/2}\,
\mathbf{e}\!\left(
\mathbf{T}_L(\hat{\mathbf{h}}_N^{\mathrm{sub}}),
\bar{\mathbf{T}}_L^{\star}
\right)\\
\mathbf{W}_R^{1/2}\,
\mathbf{e}\!\left(
\mathbf{T}_R(\hat{\mathbf{h}}_N^{\mathrm{sub}}),
\bar{\mathbf{T}}_R^{\star}
\right)\\
\sqrt{w_{\mathrm{reg}}}\,
\left(
\hat{\mathbf{h}}_N^{\mathrm{sub}}
-
\mathbf{h}_N^{\mathrm{sub}}
\right)
\end{bmatrix},
\label{eq:residual_geom}
\end{equation}
where the hand-specific weighting matrix is defined as:
\begin{equation}
\mathbf{W}_e
=
\begin{bmatrix}
w_e^{p}\mathbf{I}_3 & \mathbf{0}\\
\mathbf{0} & w_e^{R}\mathbf{I}_3
\end{bmatrix},
\qquad e\in\{L,R\}.
\label{eq:pose_weight_matrix}
\end{equation}
Here, \(w_e^{p}\) and \(w_e^{R}\) denote the translational and rotational weights for hand \(e\), respectively. \(\mathbf{h}_N^{\mathrm{sub}}\) is the terminal state of the raw trajectory, and \(w_{\mathrm{reg}}\) regularizes the solution toward the terminal state recovered from the generated video. This regularization prevents unnecessary deviations from the demonstrated motion while allowing the terminal hand poses to better satisfy the contact-aware geometric constraints.

The terminal refinement is formulated as the following constrained nonlinear least-squares problem:
\begin{equation}
\begin{aligned}
\mathbf{h}_{N}^{\mathrm{opt}}
&=
\arg\min_{\hat{\mathbf{h}}_N^{\mathrm{sub}}}
\frac{1}{2}
\left\|
\mathbf{r}_{\mathrm{geom}}(\hat{\mathbf{h}}_N^{\mathrm{sub}})
\right\|_2^2 \\
\textrm{s.t.}\quad
& z^{\min} \le z \le z^{\max},\\
& -\pi \le \psi \le \pi,\\
& \phi^{\min} \le \phi \le \phi^{\max},\\
& q_j^{\min} \le q_j \le q_j^{\max},
\qquad j=1,\dots,14.
\end{aligned}
\label{eq:terminal_ik}
\end{equation}
The optimized terminal state \(\mathbf{h}_{N}^{\mathrm{opt}}\) adjusts the root pose, waist pitch, and upper-body arm joints to align both hands with the compliance-aware object interaction targets. In practice, the resulting terminal correction is smoothly propagated to the preceding frames within a short temporal window to avoid abrupt motion changes (for details see Appendix \ref{appendix3}).



\subsection{Closed-Loop Trajectory Tracking}\label{sec:tracking}
Once the optimized trajectory is obtained, it is executed on the humanoid robot through a closed-loop tracking controller. Specifically, the robot first performs real-time localization using its onboard LiDAR to estimate its current global pose during execution. Based on this feedback, the lower body tracks the optimized global root trajectory in the Cartesian space, i.e., the root translation along the \(x\), \(y\), and \(z\) axes, while the upper body tracks the optimized joint trajectory, including the waist pitch angle and the arm joint angles of both hands. This decoupled tracking strategy allows the lower body to realize stable locomotion while enabling the upper body to maintain the desired hand-object interaction. The tracking controller is built upon the open-source Sonic general-purpose motion controller \cite{luo2025sonic}.

\begin{figure*}[t]	
	\centering
	\includegraphics[width=0.98\linewidth]{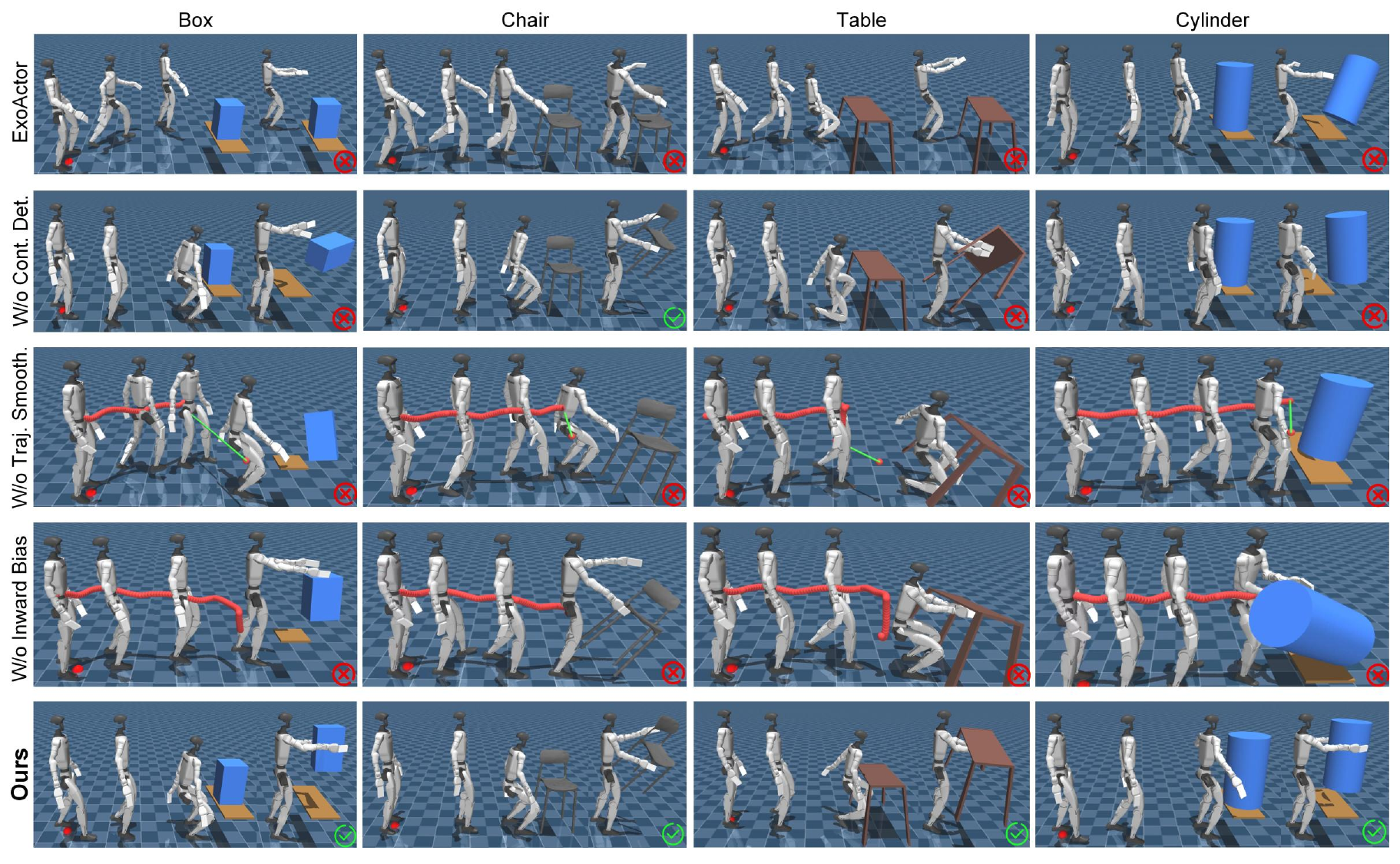}
	\caption{\textbf{Qualitative comparison of different methods on loco-manipulation tasks across four object categories.}
    Red crosses denote failed trials, and green check marks denote successful task completion.
    }
	\label{fig:3}
\end{figure*}

\begin{table*}[t]
  \centering
  \caption{Quantitative Comparison of Success Rate and Hand-Contact Point Error Across Different Objects}
  \label{tab:sim_track_error}
  \resizebox{1.95\columnwidth}{!}{
    \begin{tabular}{cccccccccccc}
    \toprule

    \multirow{2}[4]{*}{Methods} & 
    \multicolumn{5}{c}{Success Rate $\uparrow$} & 
    \multicolumn{5}{c}{Hand–Contact Point Error [m] $\downarrow$} & \\
    \cmidrule(lr){2-6} \cmidrule(lr){7-11} 
    & Box  & Chair & Table & Cylinder & Ave.
    & Box  & Chair & Table & Cylinder & Ave.\\
    \midrule
    ExoActor \cite{zhou2026exoactor}
      & 2/15 & 0/15  & 1/15 & 4/15  & 11.7\% 
      & 0.54 & 0.79 & 1.12 & 0.53 & 0.75 \\

    W/o Traj. Smooth.
     & 7/15 & 2/15 & 2/15 & 6/15 & 28.3\%  
     & 0.32 & 0.27 & 0.48 & 0.41 & 0.37\\
     
    W/o Cont. Det.
    & 5/15 & 4/15 & 9/15& 7/15 & 41.7\%  
    & 0.22 & 0.20 & 0.45 & 0.22 & 0.27\\

    W/o Inward Bias
     & 4/15 & 2/15 & 10/15 & 10/15 & 43.3\%  
     & 0.19 & 0.15 & 0.41 & \textbf{0.18} & 0.23\\
      
    \textbf{Ours}
    & \textbf{11/15} & \textbf{10/15} & \textbf{13/15 }
      & \textbf{12/15 }& \textbf{76.7\% } & \textbf{0.12} & \textbf{0.15} & \textbf{0.40} & 0.20  & \textbf{0.22}\\  

    \bottomrule
    \end{tabular}
    }
\end{table*}


\section{Experiments}
This section presents the experimental results to systematically evaluate the proposed method.
Specifically, we organize our experiments to answer the following questions:
$\mathcal{Q}_1$: How does GenHOI perform compared to other methods? (Sec. \ref{sec:compare}) 
$\mathcal{Q}_2$: What key factors influence the performance of video generation and downstream execution for HOI? (Sec. \ref{sec:elements})
$\mathcal{Q}_3$: How robust are the key modules, including the VLM-based key frame selection and hand-object contact estimation? (Sec. \ref{sec:eva_ablation}) 
$\mathcal{Q}_4$: Can the framework enable the humanoid to perform different loco-manipulation tasks in real-world scenarios? (Sec. \ref{sec:real_world})

\subsection{Experimental Setup}
We conduct both simulation and real-world experiments on the Unitree G1 humanoid robot platform. In simulation, the experiments are performed in MuJoCo, where the global poses of the robot and target objects are directly obtained from the simulator.
For real-world experiments, the humanoid robot is equipped with a Mid360 LiDAR for global relocalization \cite{bi2026alore} and an Intel RealSense D435i camera for object pose estimation \cite{wen2024foundationpose}. All algorithms run on a single workstation equipped with an Intel Core i9 3.60~GHz CPU and an NVIDIA GeForce RTX 4080 GPU. The workstation communicates with the robot over a local area network.

For the geometry-guided trajectory optimization, we use the following hyperparameters:
\(w_e^{p}=20\), \(w_e^{R}=5\), \(w_{\mathrm{reg}}=0.25\), and \(\delta=0.06\)\,m.
The bounds on the root height and waist pitch are set to
\(z^{\min}=0.3~\mathrm{m}\), \(z^{\max}=0.8~\mathrm{m}\),
\(\phi^{\min}=0.0~\mathrm{rad}\), and \(\phi^{\max}=0.58~\mathrm{rad}\), respectively.
The joint limits \(q_j^{\min}\) and \(q_j^{\max}\) are specified according to the robot URDF model.

\subsection{Comparison with Baselines}\label{sec:compare}
The most closely related works to ours are Dream2Act \cite{xu2026morphology} and ExoActor \cite{zhou2026exoactor}, both of which extract body trajectories from videos and directly track them for control. We therefore adopt \textbf{ExoActor} as a representative baseline for this line of work. To assess the importance of accurate hand--object contact localization, we introduce an ablation baseline, \textbf{W/o Cont. Det.}, which removes the contact-point detection module and instead uses the coarse 3D hand positions lifted from 2D hand masks as optimization targets. Besides, to evaluate the contributions of trajectory smoothing and the inward compressive bias, we further introduce two ablation variants, denoted as \textbf{W/o Traj. Smooth.} and \textbf{W/o Inward Bias}, respectively. For a broader comparison, we also include \textbf{HDMI} \cite{weng2025hdmi}, a representative task-specific policy-learning method. We evaluate all methods across four tasks of varying difficulty, including box grasping, asymmetric bimanual chair carrying (i.e., holding the seat with one hand while stabilizing the backrest with the other), table lifting from below, and cylindrical-object enveloping. For quantitative evaluation, we report task success rate, hand-contact point error, skill acquisition time, and generalization to out-of-distribution (OOD) object poses. Specifically, the hand-contact point error is defined as the sum of the Euclidean distances between each hand and its corresponding target contact point. To ensure a fair comparison and decouple the impact of video generation failures, all methods are evaluated on the same set of successfully generated videos.

\subsubsection{\textbf{Comparison of Success Rate and Contact Accuracy}}\label{subsec:success rate}
Fig. \ref{fig:3} presents qualitative comparisons across four object categories, with all objects initialized 2.0\,m away from the robot. For \textbf{ExoActor}, we observe significant spatial deviations between the generated trajectories and the target objects, particularly in the box grasping and table manipulation tasks. This is mainly caused by the scale ambiguity introduced when lifting 3D human motions from monocular videos, resulting in misalignment between the recovered trajectory and the real-world metric space. By contrast, the \textbf{w/o contact detection} baseline incorporates object pose information to refine the global trajectory, which effectively reduces root-level deviations and enables more accurate object approach. However, without explicit hand-object contact optimization, the recovered hand trajectories still suffer from geometric inaccuracies, leading to unreliable contact formation and failures such as object collision or unstable grasping due to excessive hand separation. The \textbf{w/o trajectory smoothing} baseline considers terminal contact optimization but does not enforce temporal smoothness on the optimized trajectory. As illustrated by the red trajectory in Fig. \ref{fig:3}, the optimized terminal pose introduces abrupt changes with respect to the preceding motion, resulting in discontinuities in the reference trajectory. These discontinuities increase tracking errors during execution and consequently degrade task success rates. This issue is particularly evident in the box grasping task, where the large terminal correction leads to significant tracking deviations. The \textbf{w/o inward bias} baseline highlights the importance of incorporating interaction-specific physical priors. Although it considers contact optimization and trajectory smoothing, stable bimanual holding still requires sufficient compressive interaction between the hands and the object. Without the inward compressive bias, the robot can achieve geometric alignment with the object but lacks the additional force-inducing prior necessary for maintaining stable contacts. In comparison, \textbf{our method} jointly addresses global trajectory alignment, contact detection, and trajectory smoothing. It achieves more precise and physically consistent object interactions.

\begin{table}[t]
  \centering
  \caption{Quantitative Comparison of Learning Time }
  \label{tab:table3}
    \begin{tabular}{cccccc}
    \toprule
    Metric & HDMI \cite{weng2025hdmi}   & ExoActor \cite{zhou2026exoactor}  & \textbf{Ours}  \\
    \midrule
    Ave. Learning Time
    & $\sim$70\,min & 1\,min\,33\,s & 1\,min 51\,s   \\

    \bottomrule
    \end{tabular}
\end{table}

\begin{table}[t]
  \centering
  \caption{Comparison of Success Rates under OOD Robot-Object Distance}
  \label{tab:table4}
  \resizebox{0.9\columnwidth}{!}{
    \begin{tabular}{cccccc}
    \toprule
     \multirow{2}[4]{*}{Methods} & 
    \multicolumn{5}{c}{Robot-Object Distance [m]} \\
    \cmidrule(lr){2-6}
      & -1.0 & -0.5 & +0.5 & +1.0 & +1.5 \\
    \midrule
    HDMI \cite{weng2025hdmi} 
      & 3/10 & 8/10 & 8/10 
      & 4/10   & 0/10\\
    ExoActor \cite{zhou2026exoactor}
    & 0/10 & 0/10 & 0/10 
      & 0/10  & 0/10 \\
    \textbf{Ours}
    & \textbf{8/10}& \textbf{10/10} & \textbf{8/10} 
      & \textbf{7/10}   & \textbf{8/10} \\
    \bottomrule
    \end{tabular}
  }
\end{table}

To quantitatively evaluate performance, we report task success rate and hand-contact point error across four object categories, as shown in Table \ref{tab:sim_track_error}. For each category, objects are initialized at five positions in front of the robot, ranging from 1.0\,m to 2.0\,m with 0.25\,m intervals, and one video is generated per position. Each trial is repeated three times, resulting in 15 evaluations per category. A trial is considered successful if the object is lifted above the supporting surface for at least 1\,s.
The results show that our method consistently outperforms both baselines, achieving the highest average success rate of 76.7\% and the lowest average hand--contact point error of 0.22\,m. Similar to the analysis above, most failure cases can be attributed to root-position tracking errors. These errors propagate to the hand trajectories, increasing the hand--contact point error and occasionally causing unintended collisions with the object. ExoActor suffers from global geometric inconsistencies inherited from direct video-to-trajectory extraction, leading to systematic directional deviations and unreliable contact formation. The baseline without contact-point detection performs better than ExoActor, as its object-aware refinement partially alleviates root-trajectory errors; however, the lack of explicit contact localization still limits stable object interaction due to inaccurate hand-object alignment. The w/o trajectory smoothing baseline shows that abrupt trajectory corrections degrade tracking accuracy, which further increases the hand-contact point error and leads to lower task success rates. In addition, although the w/o inward bias baseline uses the smoothed trajectory, it still relies on purely geometry-based tracking and lacks an explicit prior for inducing object-holding forces. As a result, the robot may achieve accurate geometric alignment but fail to establish sufficiently stable contacts for holding the object. As a result, all these variants achieve success rates below 50\%, accompanied by the average hand-contact point errors of up to 0.75\,m.

\begin{figure}[t]	
	\centering
	\includegraphics[width=0.95\linewidth]{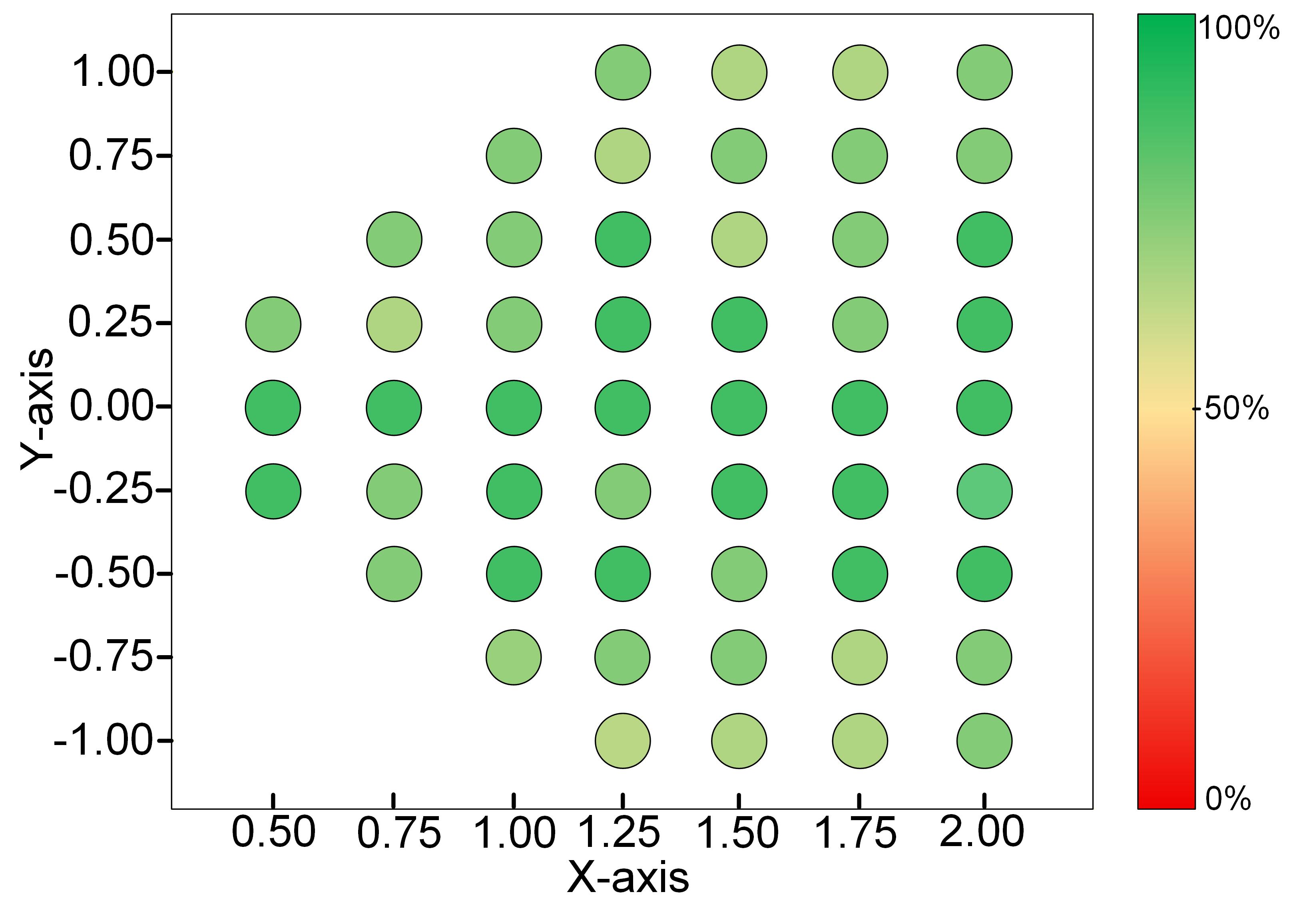}
	\setlength{\abovecaptionskip}{-5pt} 
	\caption{\textbf{Task success rates under OOD box positions.}}
	\label{fig:4}
\end{figure}

\subsubsection{\textbf{Comparison of Learning Time}}\label{subsec:time}
In addition, task-specific training methods such as HDMI tend to overfit a single trajectory, achieving near 100\% success rates \cite{weng2025hdmi}. However, these methods require object-specific policy training and are computationally expensive, while exhibiting limited generalization to novel object configurations.
Table \ref{tab:table3} reports the average learning time on the box carrying task over three runs for each method. To ensure a fair comparison, all methods are evaluated under identical robot-object relative poses and on the same hardware platform, namely a single workstation equipped with an Intel Core i9 3.60 GHz CPU and an NVIDIA GeForce RTX 4080 GPU.
The results show that HDMI requires substantially longer training time than other methods, taking more than one hour to complete a single task. In contrast, our method requires less than 2 minutes while still achieving a relatively high success rate, demonstrating significantly improved efficiency.

\begin{figure*}[t]	
	\centering
    \includegraphics[width=0.95\linewidth]{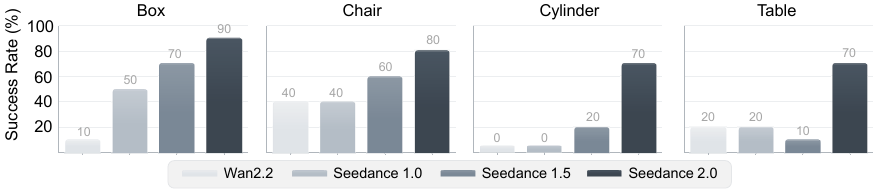}
	\caption{\textbf{Success rates of video generation models across four manipulation tasks.} Each bar reports the generation success rate over 10 generated videos.
    }
	\label{fig:5}
\end{figure*}

\subsubsection{\textbf{Comparison of Generalization Performance}}\label{subsec:OOD}
We evaluate the generalization ability of different methods under OOD object poses using the box-carrying task. The reference trajectory is generated with the box initialized 1.5\,m from the robot, while evaluation is conducted at unseen distances of 0.5\,m, 1.0\,m, 2.0\,m, 2.5\,m, and 3.0\,m. Each distance is evaluated over 10 trials, with the results summarized in Table \ref{tab:table4}.
Our method consistently outperforms the baselines across OOD object distances. This is mainly because our approach refines the robot trajectory conditioned on the current object pose, rather than simply tracking a fixed reference trajectory.
In contrast, HDMI learns to track a fixed trajectory and remains effective when test poses are close to the training distribution. However, its performance degrades significantly once the object is placed outside the randomized range used during training, reaching 0\% success at 3.0\,m.
Similarly, ExoActor achieves lower success rates, as it directly tracks an open-loop trajectory recovered from video and cannot steer the robot toward unseen object poses.

To further evaluate the generalization capability of the proposed method, we conduct the box-carrying task under OOD box positions using a single reference trajectory. As shown in Fig. \ref{fig:4}, each point reports the success rate averaged over 10 repeated trials at a specific box position. Some positions do not have recorded results because they fall outside the field of view (FOV) of the robot-mounted camera.
The results show that the proposed method consistently achieves high success rates across a wide range of OOD positions (greater than 65\%), owing to its decoupled design that enables online trajectory refinement conditioned on the observed object pose.

\begin{figure}[t]	
	\centering
	\includegraphics[width=0.95\linewidth]{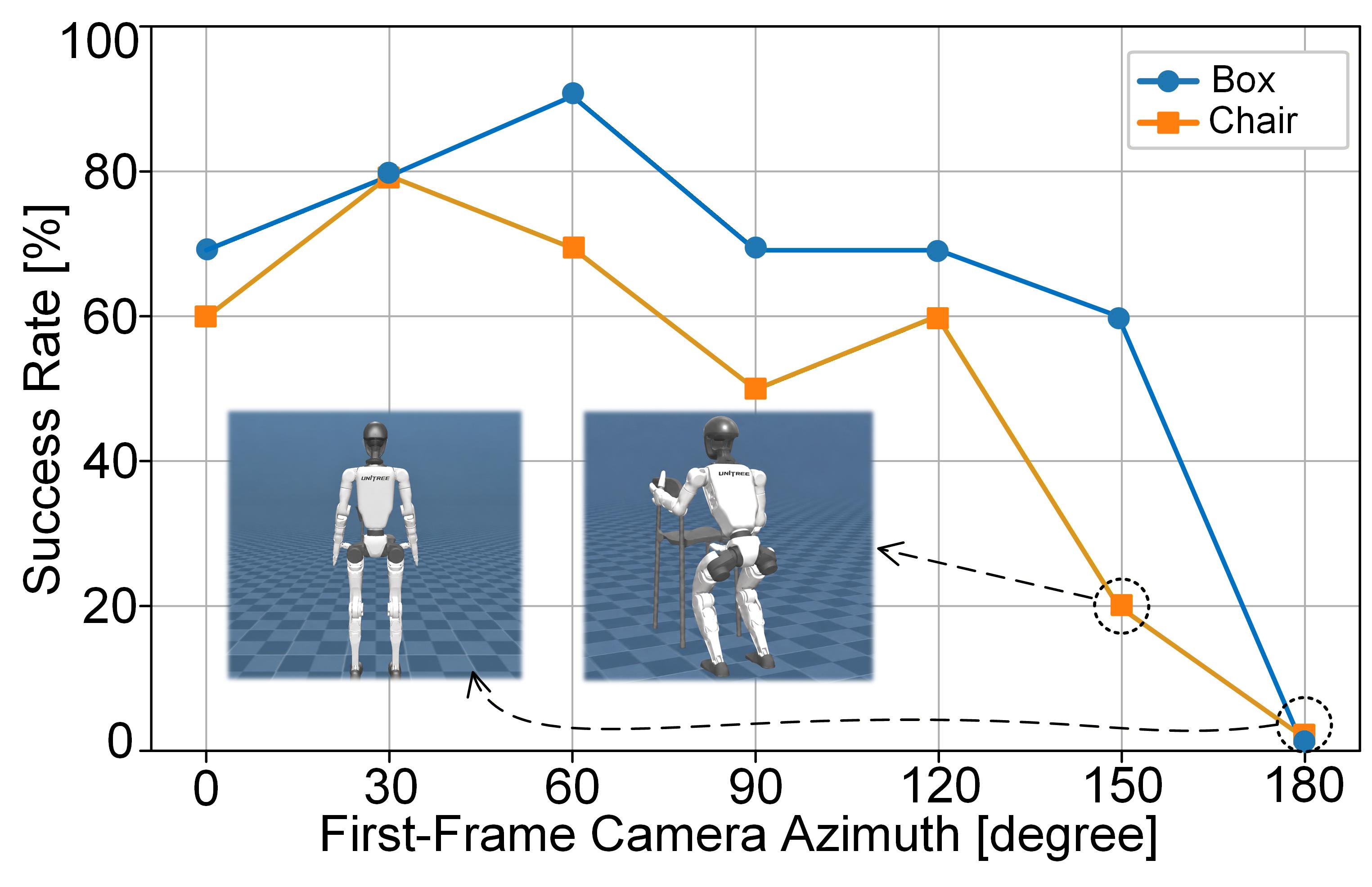}
	\setlength{\abovecaptionskip}{-5pt} 
	\caption{\textbf{Effect of first-frame camera azimuth on task success rates. }Success rates for the box and chair tasks are evaluated at azimuth angles from \(0^\circ\) to \(180^\circ\), with representative generated frames shown for selected viewpoints.
    }
	\label{fig:6}
\end{figure}

\subsection{Key Factors  of Video Generation for Loco-Manipulation}\label{sec:elements}
To understand how video generation influences the downstream performance of GenHOI in simulation and real-world deployment, we study two controllable factors in the generation process: the video generation model and the camera azimuth used to capture the first-frame image.

\subsubsection{\textbf{Video Generation Model}}
To evaluate the impact of different video generation models on the quality of generated task-oriented videos, we compare Wan 2.2, Seedance 1.0, Seedance 1.5, and Seedance 2.0 across four loco-manipulation tasks involving a box, chair, table, and cylinder. For each model-task pair, we generate 10 videos and evaluate whether each video satisfies the generation criteria, including faithful instruction following, consistent object geometry, and fixed camera viewpoint. The results are summarized in Fig.~\ref{fig:5}.
Overall, stronger video generation models, such as Seedance 2.0, achieve higher video generation success rates by producing videos with better prompt adherence, more consistent object geometry, and more physically plausible interactions. In contrast, Wan 2.2 and Seedance 1.0 frequently suffer from object deformation and camera-viewpoint drift, resulting in invalid task-oriented videos. For table and cylinder manipulation, video generation is particularly challenging due to more complex hand-object interactions, where physically inconsistent behaviors, such as object deformation, are more likely to occur.

\subsubsection{\textbf{First-Frame Camera Azimuth}}
To analyze the effect of the first-frame camera viewpoint, we vary the camera azimuth around the robot–object scene for both the box and chair tasks, while keeping other camera parameters fixed, including the camera height, pitch angle, and so on. Due to the bilateral symmetry of the robot-object setup, we only evaluate azimuth angles from \(0^\circ\) to \(180^\circ\) at \(30^\circ\) intervals. For each angle, we render one first-frame image and generate five valid videos using Seedance 2.0. Each generated video is processed through GenHOI twice to execute the task. The task success rates are reported in Fig. \ref{fig:6}.
The results indicate a noticeable drop in success rate at azimuth angles of \(150^\circ\) and \(180^\circ\). This is because, from these viewpoints, the robot substantially occludes the target object in the generated videos, leading to unreliable hand detection and contact-point estimation. In contrast, for viewpoints where both the robot and the target object are clearly visible, the proposed method maintains consistent performance across different camera azimuths.

\subsection{Key Modules Evaluation}\label{sec:eva_ablation}

\subsubsection{\textbf{Evaluation of Key Frame Selection}}
To evaluate the accuracy of VLM-based key frame selection, we test two vision-language models, Doubao-Seed-2.0 and GPT-5.5, on four object manipulation tasks: box, chair, table, and cylinder. For each object, we generate five videos and query the VLM three times for each video. A selection is counted as successful if the predicted frame correctly captures the initial hand-object contact. The results are reported in Table~\ref{tab:table5}.

Both models achieve high selection accuracy across different objects, with average accuracies of at least \textbf{95\%}, indicating that VLMs can reliably localize the critical interaction moment from generated videos. The only noticeable performance drop occurs in the table task, where the larger object geometry and partial hand-object occlusions make the first contact moment more ambiguous. Nevertheless, the consistently high accuracy demonstrates that the proposed key frame selection module provides reliable temporal anchors for subsequent contact estimation and motion extraction.

\subsubsection{\textbf{Evaluation of Contact Point Detection}}
\begin{figure}[t]	
	\centering
	\includegraphics[width=0.9\linewidth]{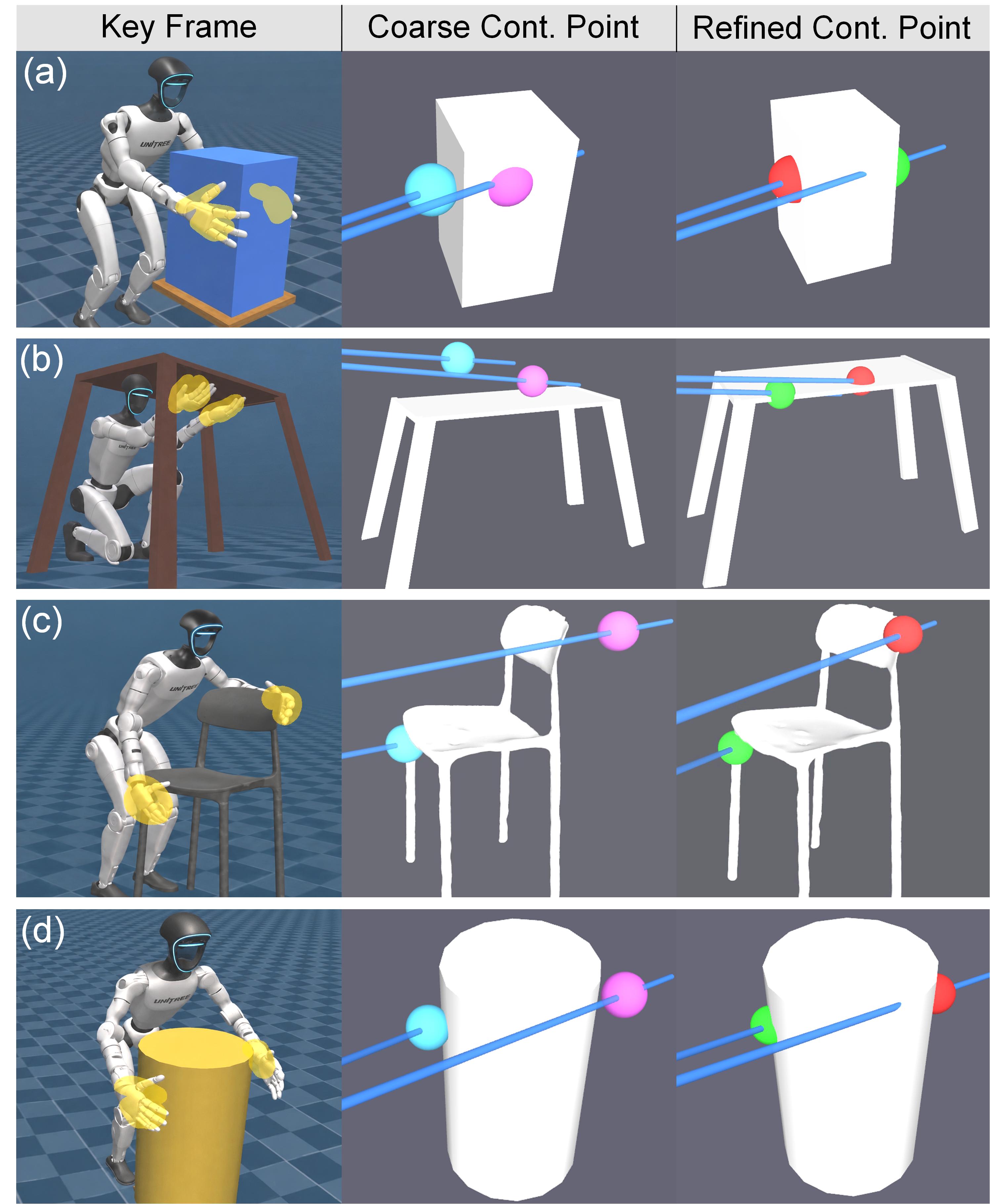}
	\caption{\textbf{Visualization of contact point detection.} Each row corresponds to one manipulation task, and the three columns show the selected key frame, coarse contact points, and refined contact points, respectively.
    }
	\label{fig:7}
\end{figure}

Fig.~\ref{fig:7} illustrates contact point detection on different object manipulation tasks. For each task, the first column shows the selected key frame with detected hand masks. The second column visualizes the coarse contact points, which are obtained by back-projecting the depth pixels within the hand masks into 3D using the estimated depth map and camera intrinsics. The refined contact points are shown in the third column.
The results show that, even when the hands are partially occluded or only weakly visible in the key frame, the proposed refinement step effectively recovers reliable hand-object contact locations.

\begin{table}[t]
  \centering
  \caption{Evaluation of VLM-Based Key Frame Selection}
  \label{tab:table5}
    \begin{tabular}{cccccccc}
    \toprule
    Models & Box & Chair  & Table  & Cylinder & Ave.\\
    \midrule
    Doubao-Seed-2.0

    & 15/15 & 15/15 & 12/15 & 15/15 & 95.0\%   \\

     GPT-5.5
    & 15/15 & 15/15 & 13/15 & 15/15 & 96.7\%    \\

    \bottomrule
    \end{tabular}
\end{table}

\begin{figure*}[t]	
	\centering
	\includegraphics[width=0.95\linewidth]{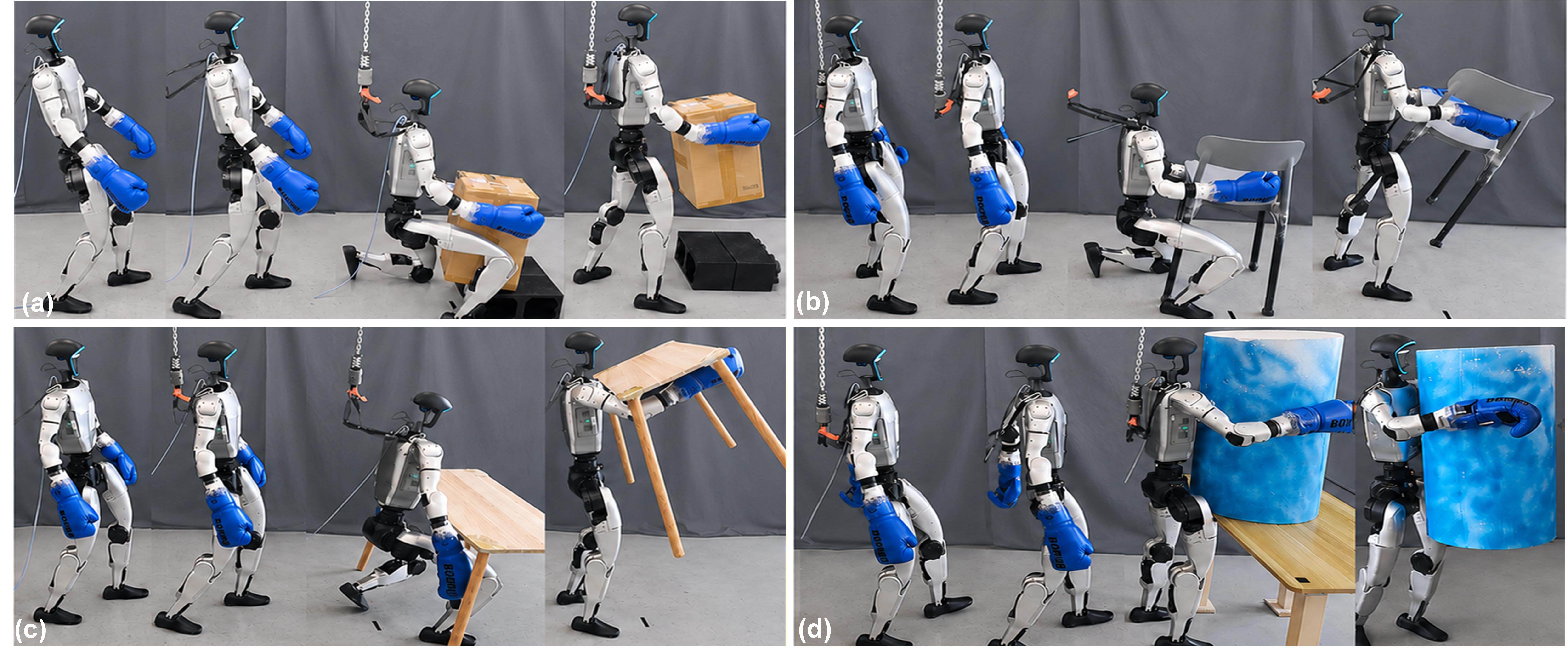}
	\setlength{\abovecaptionskip}{-5pt} 
	\caption{\textbf{Real-world experiments across four object-interaction tasks.} Including box grasping (a), asymmetric bimanual chair carrying (b), table lifting from below (c), and cylindrical-object enveloping (d).}
	\label{fig:real_1}
    \vspace{-10pt}
\end{figure*}

\begin{figure*}[t]	
	\centering
	\includegraphics[width=0.95\linewidth]{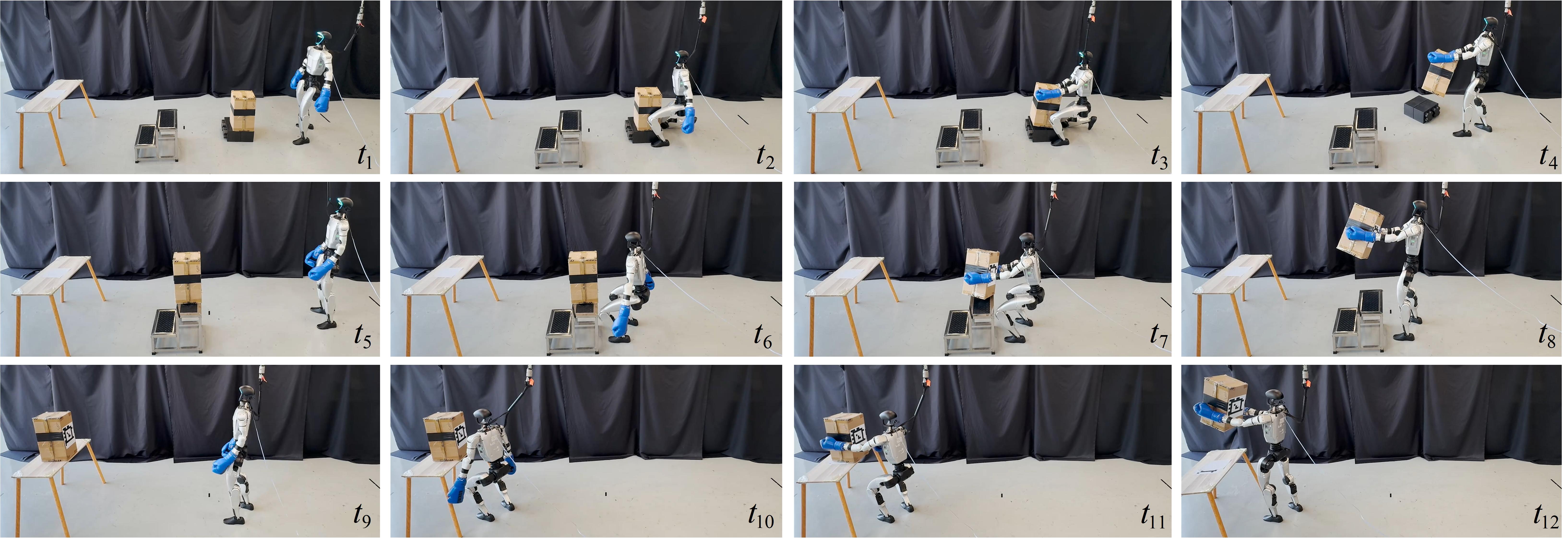}
	\setlength{\abovecaptionskip}{-0pt} 
	\caption{\textbf{Continuous real-world box-grasping demonstrations across OOD object poses.}}
	\label{fig:real_2}
    \vspace{-10pt}
\end{figure*}

\subsection{Real-World Experiments}
\label{sec:real_world}
We conduct real-world experiments to evaluate the proposed GenHOI framework. The experiments include box grasping, asymmetric bimanual chair carrying, table lifting from below, and cylindrical-object enveloping, as well as continuous box grasping at varying positions.
These experiments aim to:
(1) demonstrate the ability of our method to rapidly acquire object-interaction skills across diverse objects;
(2) highlight the steerable interaction capability, where a single reference trajectory can generalize to objects at different positions.

\subsubsection{\textbf{Interaction with Different Objects}}\label{sec:diff_obj}
As illustrated in Fig.~\ref{fig:real_1}, we first present qualitative demonstrations to evaluate whether GenHOI can achieve loco-manipulation across different objects. In each task, the target object is placed 1.0--3.0\,m away from the robot, ensuring that its initial pose remains within the camera FOV. Starting from the perceived object pose, the robot reconstructs the scene in simulation, generates a video conditioned on the rendered first-frame image and a language command, extracts geometric constraints, and executes the optimized trajectory with a closed-loop tracking controller.
Specifically, we evaluate four representative objects with different geometries and contact configurations: a box ($0.23\,\mathrm{m}\times0.26\,\mathrm{m}\times0.42\,\mathrm{m}$), a chair ($0.37\,\mathrm{m}\times0.37\,\mathrm{m}\times0.85\,\mathrm{m}$), a table ($0.80\,\mathrm{m}\times0.36\,\mathrm{m}\times0.60\,\mathrm{m}$), and a cylinder ($\Phi\,0.40\,\mathrm{m}\times0.70\,\mathrm{m}$). The box is initialized 1.8\,m in front of the robot with its centroid 0.3\,m above the ground. The chair is placed on the ground at 1.6\,m, the table is placed on the ground at 1.5\,m, and the cylinder is initialized at 1.5\,m with its centroid 0.9\,m above the ground. GenHOI generates distinct interaction strategies for these objects: the robot establishes symmetric bimanual contacts to lift the box, forms asymmetric contacts on the chair backrest and seat, bends down to lift the table from underneath, and adopts an enveloping motion around the cylinder. These results show that our method can adapt the generated motion to diverse object geometries and contact configurations.

\subsubsection{\textbf{Generalization to Out-of-Distribution Object Positions}}\label{sec:OOD}
As shown in Fig. \ref{fig:real_2}, the box is placed at three different planar positions with distances of 1.5\,m, 2.5\,m, and 3.5\,m. Using only a single box-grasping trajectory extracted from the generated video, the robot continuously performs grasping tasks across these varying object configurations. During $t_1$--$t_4$, the robot approaches and grasps a box placed at a height of 0.3\,m above the ground while adopting a single-knee posture. The box is then relocated to a platform at a height of 0.6\,m. During $t_5$--$t_8$, the robot re-estimates the box pose, optimizes the reference trajectory based on the current object configuration, and grasps the box in a semi-squatting posture. Finally, the box is placed on a table at a height of 0.8\,m. From $t_9$ to $t_{12}$, the robot again adapts the trajectory according to the real-time box pose and successfully completes the grasping task. Notably, all three sequential tasks are completed using only a single generated video and the corresponding reference trajectory extracted from it, with online trajectory optimization enabling adaptation to different OOD object heights and positions.

\subsection{Analysis of Failure Cases}\label{sec:failure}
\begin{figure}[t]	
	\centering
	\includegraphics[width=0.95\linewidth]{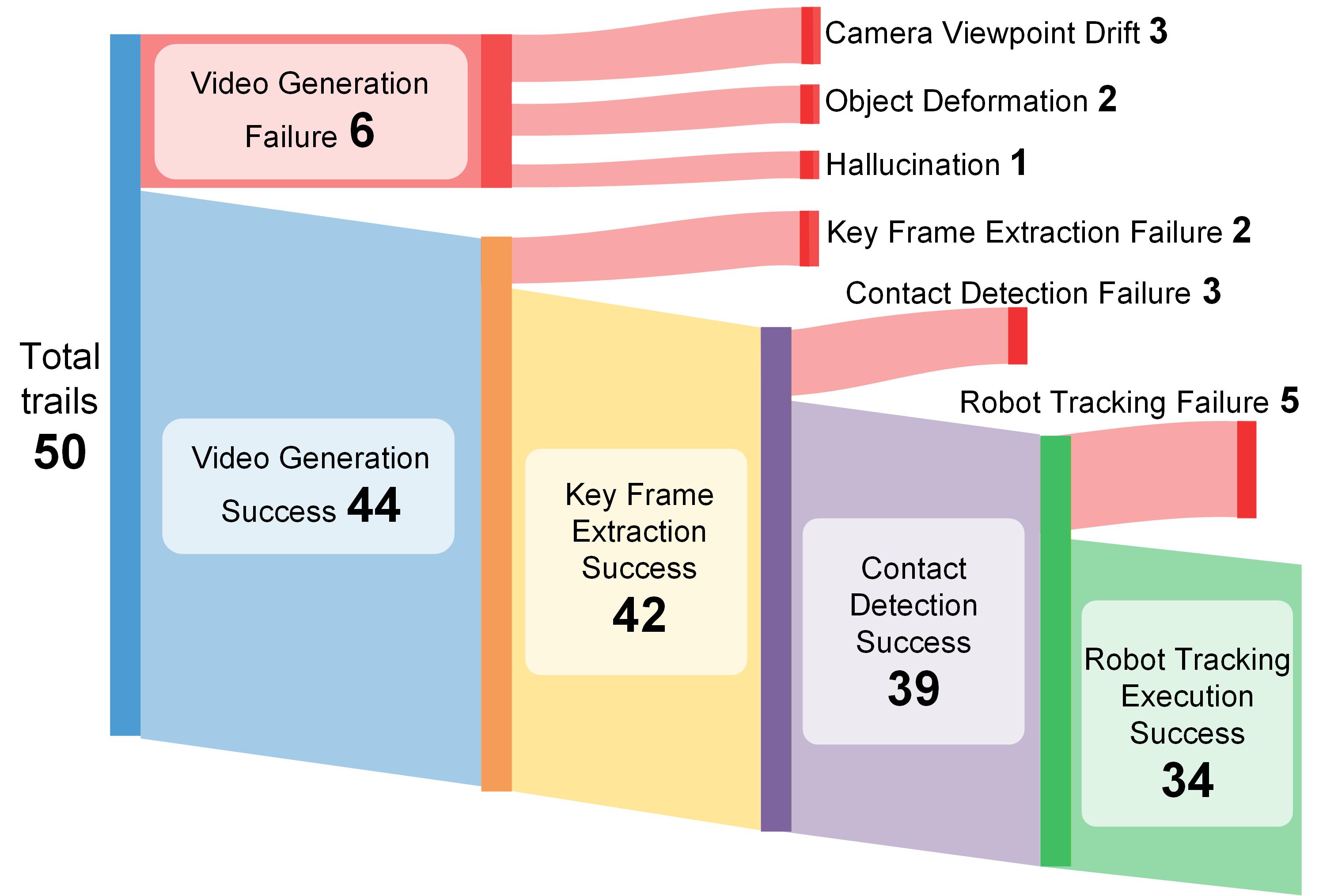}
	\caption{\textbf{Failure-case analysis of the framework.} Across video generation, key frame extraction, contact detection, and trajectory tracking.
    }
	\label{fig:10}
    \vspace{-10pt}
\end{figure}

Since GenHOI consists of multiple sequential modules, failures may originate from different stages and propagate through the pipeline. To better understand the contribution of each module to the final task outcome, we conduct 50 repeated trials on the box-grasping task and categorize the failure cases, as shown in Fig.~\ref{fig:10}. Overall, 34 out of 50 trials successfully complete the full pipeline. The main failures occur during video generation, including camera-viewpoint drift, object deformation, and hallucination, followed by errors in keyframe extraction and contact detection. In addition, five trials fail during robot trajectory tracking, mainly due to the limited tracking accuracy of the current low-level humanoid controller. Overall, directly imitating generated videos for HOI is primarily limited by the quality of the generated videos and the tracking accuracy of the humanoid locomotion controller.

\section{Conclusion}
In this paper, we presented GenHOI, a framework that enables humanoid robots to perform object-interaction tasks by imitating a single generated video. Given an onboard observation and a language command, GenHOI reconstructs the robot--object scene in simulation, generates a task video, extracts contact-aware geometric constraints, and refines the recovered motion through geometry-guided trajectory optimization. The optimized trajectory is then executed by a closed-loop whole-body controller, enabling real-world loco-manipulation without task-specific policy training or physical demonstration data.
Simulation and real-world experiments on a Unitree G1 robot demonstrate diverse interaction skills, including box grasping, table lifting, chair carrying, and cylindrical-object enveloping. The results further show that GenHOI enables steerable interactions and generalizes to OOD object poses. Overall, this work represents a step toward using generative video models as scalable action priors for HOI.

The current framework has several limitations. First, the real-to-sim reconstruction assumes access to an accurate object mesh, which may limit its scalability to previously unseen objects. Future work could incorporate online 3D reconstruction or shape completion from onboard RGB-D observations. Second, the current platform lacks dexterous hands, restricting the system to relatively coarse interactions. Future work incorporates dexterous end-effectors to support more precise and versatile humanoid–object interaction.

\bibliographystyle{IEEEtran}
\normalem
\bibliography{ref}


\appendices
\input{appendix}

\end{document}

%% file: appendix.tex
\section{Video Generation Prompts}\label{appendix1}
\begin{figure}[t]	
	\centering
\includegraphics[width=1.0\linewidth]{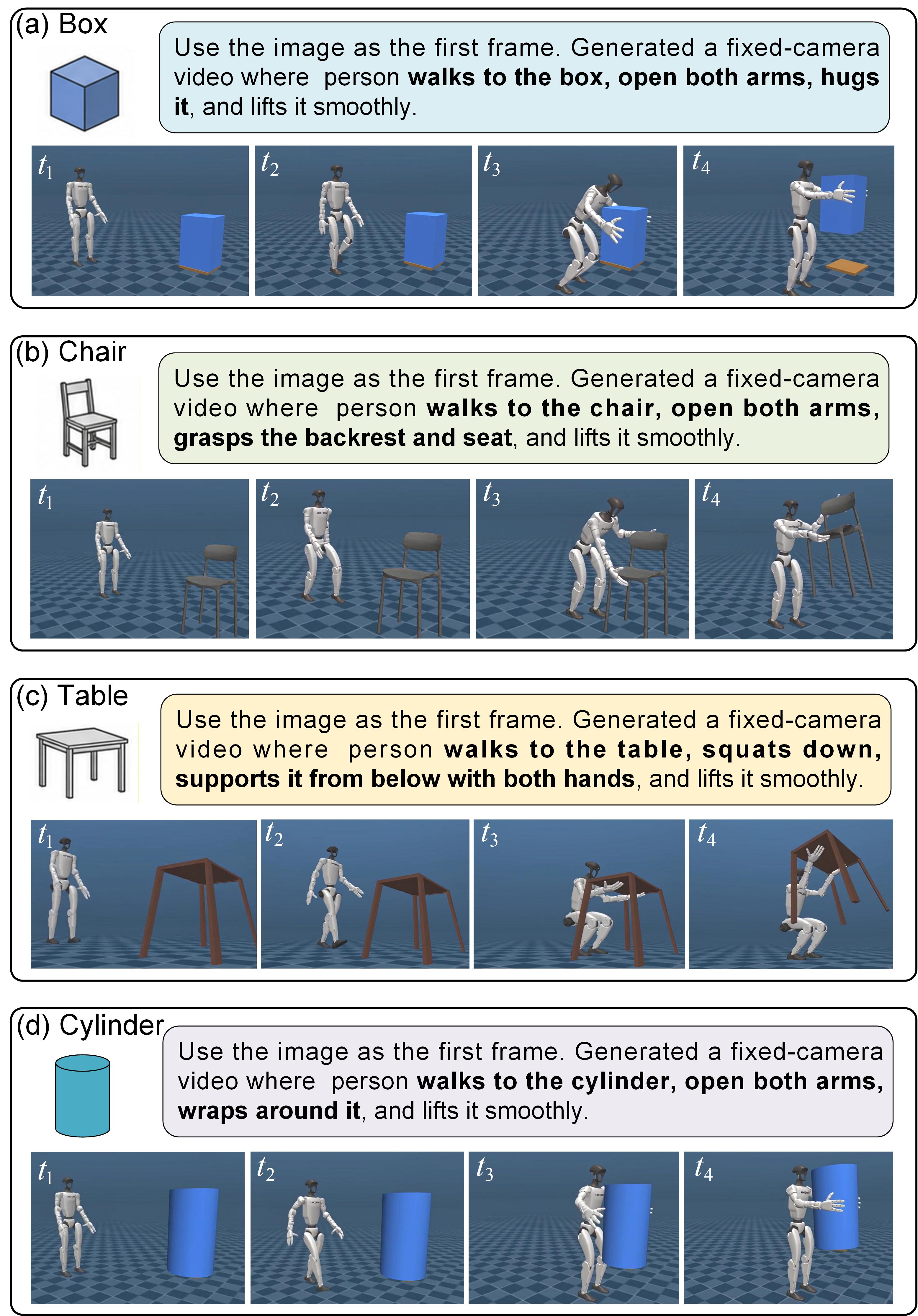}
	\caption{Object-specific prompts and representative key frames from the generated videos.  
    }
	\label{fig:AP1}
\end{figure}

To guide the video diffusion model toward physically plausible and task-consistent outputs, we design object-specific text prompts for each loco-manipulation task. 
Fig.~\ref{fig:AP1} shows the prompts used for four representative objects, including a box, chair, table, and cylinder, together with four selected key frames from the generated videos.

\section{Key Frame Selection}\label{appendix2}

\begin{figure}[t]	
	\centering
\includegraphics[width=1.0\linewidth]{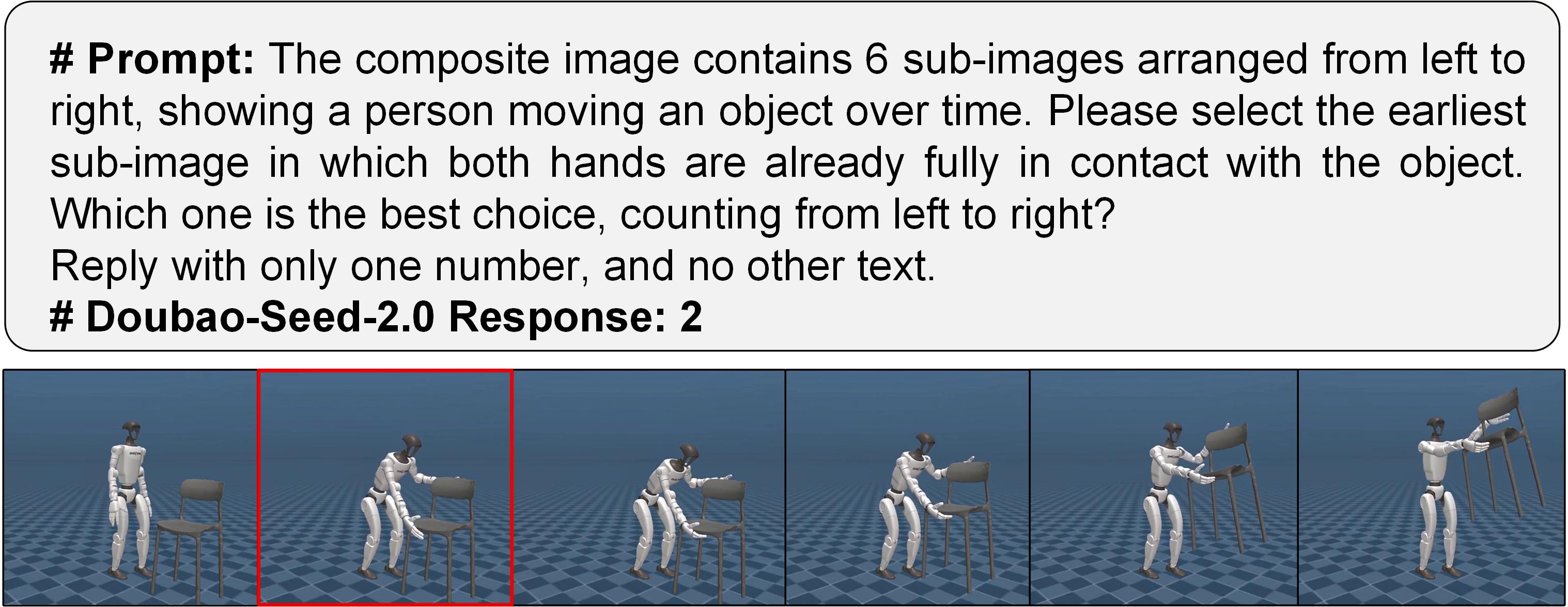}
	\caption{Prompt of the key frame selection. 
    }
	\label{fig:AP2}
\end{figure}

Fig.~\ref{fig:AP2} illustrates the procedure for selecting the key frame from a generated video. We first sample a set of candidate frames from the video and concatenate them into a single composite image in chronological order. The composite image is then fed into Doubao-Seed-2.0, together with a prompt asking the model to identify the earliest frame in which both hands are fully in contact with the object. The model is required to return only the index of the selected frame, which is used as the key frame for subsequent contact-point detection and object-aware constraint extraction. In the example shown in Fig.~\ref{fig:AP2}, the second frame is selected as the key frame.

\section{Trajectory Smoothing}\label{appendix3}
After obtaining the optimized terminal state \(\mathbf{h}_N^{\mathrm{opt}}\), we compute the terminal correction relative to the reference trajectory as:
\begin{equation}
\Delta \mathbf{h}
=
\mathbf{h}_N^{\mathrm{opt}} - \mathbf{h}_N^{\mathrm{sub}}.
\end{equation}
Rather than re-optimizing the entire trajectory, we smoothly propagate this terminal correction over the last \(K\) frames of the reduced-state trajectory, where \(K=90\), corresponding to the final 3\,s of motion. Specifically, for each frame \(i \in \{N-K+1,\dots,N\}\), we define a normalized tail coordinate
\begin{equation}
s_i = \frac{i-(N-K+1)}{K-1},
\qquad s_i \in [0,1],
\end{equation}
and compute a quintic smoothstep weight:
\begin{equation}
\alpha_i = 10s_i^3 - 15s_i^4 + 6s_i^5.
\end{equation}
The refined reduced state is then updated as:
\begin{equation}
\tilde{\mathbf{h}}_i
=
\mathbf{h}_i^{\mathrm{sub}} + \alpha_i \Delta \mathbf{h},
\qquad i=N-K+1,\dots,N.
\end{equation}
Each updated state is subsequently projected onto the same box constraints used in \eqref{eq:terminal_ik} to ensure kinematic feasibility. In this way, the terminal correction is distributed smoothly over the final 3\,s of the trajectory, leading to a gradual transition to the optimized terminal configuration while avoiding abrupt motion changes.